\documentclass[a4paper]{article}
\usepackage{RR}
\usepackage{hyperref}
\usepackage{color}
\usepackage{amsmath,pifont,amssymb}
\usepackage{graphics,array,float,epsfig}
\usepackage{amsfonts,graphicx,color}
\usepackage{eclbkbox}
\usepackage{multirow}
\def\squarebox#1{\hbox to #1{\hfill\vbox to #1{\vfill}}}
\newcommand{\qed}{\hspace*{\fill}
        \vbox{\hrule\hbox{\vrule\squarebox{.667em}\vrule}\hrule}\smallskip}

\RRdate{Septembre 2007}

\RRauthor{ Patrick H\'eas
  \and
Mihai Datcu
}
\authorhead{H\'eas \& Datcu.}
\RRtitle{Apprentissage supervis\'e sur des graphes de similarit\'e spatio-temporelle dans les s\'equences d'images satellites}
\RRetitle{Supervised learning on graphs of spatio-temporal similarity in satellite image sequences}

\RRresume{
Les s\'equences d'images satellites de haute r\'esolution sont des signaux multidimensionnels compos\'es de motifs spatio-temporels associ\'es \`a des ph\'enom\`enes nombreux et vari\'es. Des m\'ethodes bay\'esiennes ont \'et\'e pr\'ec\'edemment propos\'ees dans~\cite{Heas(2005)} pour coder l'information contenue dans les s\'equences d'image satellitaire sous forme de graphes. Bas\'e sur une telle repr\'esentation, ce papier pr\'esente une m\'ethode d'apprentissage supervis\'e  de s\'emantiques associ\'ees aux motifs spatio-temporels de ces s\'equences d'images. Cela permet la reconnaissance et la recherche probabiliste de ph\'enom\`enes similaires.
En effet, les graphes repr\'esentent des mod\`eles statistiques de processus spatio-temporels, qui permettent de d\'ecrire des changements physiques observ\'es dans la sc\`ene. En cons\'equence, par apprentissage supervis\'e, un mod\`ele param\'etrique \'evaluant les types de similarit\'e entre motifs de graphes est ajust\'e pour repr\'esenter les s\'emantiques  associ\'ees \`a ces ph\'enom\`enes spatio-temporels. L'apprentissage est effectu\'e par la d\'efinition incr\'ementale de types de similarit\'es via des exemples fournis par l'utilsateur de motifs associ\'es \`a des s\'emantiques positives ou/et n\'egatives. A partir de ces exemples, des probabilit\'es sont d\'eduites par l'utilisation d'un r\'eseau bay\'esien et d'un mod\`ele de Dirichlet. Ces probabilités permettent de relier l'int\'er\^et de l'utilisateur \`a un mod\`ele de similarit\'e sp\'ecifique entre motifs de graphe. A chaque stade d'apprentissage, les probabilit\'es \textit{a posteriori} sont actualis\'ees pour l'ensemble des motifs  de graphe possibles afin que les ph\'enom\`enes spatio-temporels puissent \^etre reconnus et retrouv\'es dans la s\'equence d'image. Quelques exp\'eriences efffectu\'ees sur une s\'equence multi-spectral d'images SPOT illustrent la m\'ethode de reconnaissance spatio-temporelle propos\'ee.
}
\RRabstract{
High resolution satellite image sequences are multidimensional signals composed of  spatio-temporal patterns associated to numerous and various phenomena. Bayesian methods have been previously proposed in~\cite{Heas(2005)} to code the information contained in satellite image sequences in a graph representation using Bayesian methods. Based on such a representation, this paper further presents a supervised learning methodology of semantics associated to spatio-temporal patterns occurring in satellite image sequences. It enables the recognition and the probabilistic retrieval of similar events.
Indeed, graphs are attached to statistical models for spatio-temporal processes, which at their turn describe physical changes in the observed scene. Therefore, we adjust a parametric model evaluating similarity types between graph patterns in order to represent user-specific semantics attached to spatio-temporal phenomena. The learning step is performed by the incremental definition of similarity types via user-provided spatio-temporal pattern examples attached to positive or/and negative semantics. From these examples, probabilities are inferred using a Bayesian network and a Dirichlet model. This enables to links user interest to a specific similarity model between graph patterns. According to the current state of learning, semantic posterior probabilities are updated for all possible graph patterns so that similar spatio-temporal phenomena can be recognized and retrieved from the image sequence. Few experiments performed on a multi-spectral SPOT image sequence illustrate the proposed spatio-temporal recognition method.
}
\RRmotcle{Reconnaissance de forme,  apprentissage supervis\'e, ph\'enom\`enes spatio-temporels; similarit\'e de graphes; r\'eseaux bay\'esiens; mod\`ele de Dirichlet}
\RRkeyword{Pattern recognition; supervised learning, spatio-temporal phenomena, graph similarity; bayesian networks; Dirichlet model}
\RRprojets{VISTA}
\RRtheme{\THCom \THCog \THSym \THNum \THBio} 
\URRennes  
\begin{document}
\makeRR   

\section{Introduction}
During the last decades, the imaging satellite sensors have acquired huge quantities of data enabling the elaboration of satellite image sequences. However, our capability to store large volume of data has highly exceeded our capability to extract and interpret the relevant information. Therefore, satellite image sequences information learning systems are needed to bridge the semantic gap between information extracted from temporal and pictural multidimensional data, and user-specific interests. Indeed, satellite image sequences are complex objects possessing a rich information content. They contain numerous and various spatio-temporal structures. For example in rural scenes, one can observe the growth and the maturation of cultures, their harvests, evolutions of ploughland, river floods, etc. Near urban areas, car and plane occlusions are frequent but there are also evolving constructions, pollution phenomenon, etc.  Spatio-temporal analyses are useful to understand complex evolutions which concern various domains such as agriculture, forest monitoring, ecology, hydrology, urbanization, etc.\\

\noindent
Experiments presented in this paper were performed using a satellite image sequence composed of  SPOT multispectral images containing 2000x3000 pixels. The spatial resolution is 20 meters. The acquired scene is a rural area located in the East of Bucharest (Romania). The acquisition campaign was driven in order to provide remote sensing data for the  \textit{Data Assimilation for Agro-Modeling (ADAM)} project. The sequence was obtained by daily acquisition and by filtering out images presenting  a cloud or a snow cover above the project test sites. This selection procedure resulted in 38 images irregularly sampled in time, which were acquired over a period of 286 days. The images were then made superposable and a radiative transfer model was applied to produce reflectance measurements. The ADAM project satellite image sequence is available on-line~\cite{BDADAM}.\\

\noindent
To exploit satellite image sequence information content,  in previous work an information flow between satellite image sequences content and user interest has been established by modeling hierarchically the information content in satellite image sequences~\cite{Heas(2005)}. On the first levels of the hierarchical modeling, strong families of models are applied to extract information using inference based on Bayesian and entropic methods. This unsupervised modeling results in a graph representation coding the information content of satellite image sequences. More precisely, the modeling of the time-evolution of the distribution of features extracted  at consecutive times from the image sequence has been proposed. The modeling has resulted in a set of cluster trajectories, possibly splitting and merging in time, which are grouped into a graph $\mathcal{G}$.\\

\noindent
Based on this  objective graphical signal characterization, we focus in this paper on a very important step which is  providing content-based query techniques : the interaction with the user and the flexible incorporation of user-specific interests. This constitutes the last level of the global hierarchical information modeling introduced in~\cite{Heas(2005)}. However, Bayesian learning of similarity between graph patterns which is the kernel of this last inference level is not presented in the latter article. Therefore, the aim of the present paper is to describe this learning methodology employing examples of spatio-temporal processes provided on-line by the user.\\ 

\begin{figure}\begin{center}
\begin{tabular}{c}
\includegraphics[width=0.45\textwidth]{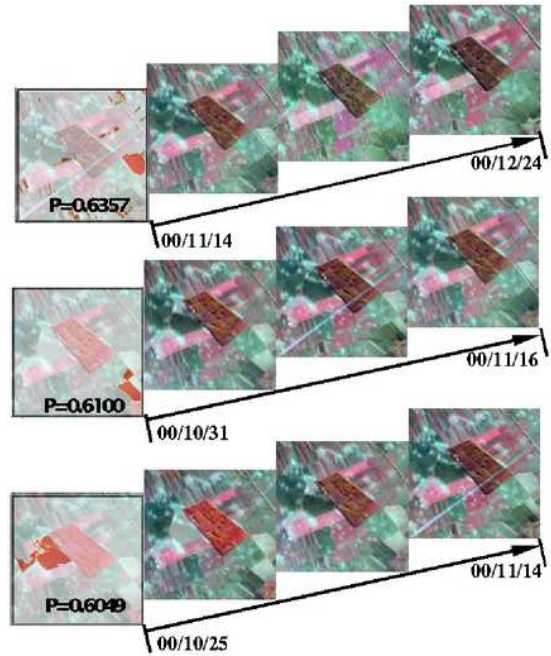}
\end{tabular}\end{center}
\caption{\small{Results of a probabilistic search  of spatio-temporal patterns possessing plowing semantics  retrieved in space (red class) and time (period written under the 3 image sequences) within a satellite image sequence.}}  
\label{fig1}
\end{figure}

\noindent
The goal of such an supervised learning procedure is the inference of  similarity measurements between the spatio-temporal processes present in the image sequences, which can then enable the retrieval of phenomena in space and time. Indeed, spatio-temporal processes present in a given time and spatial window of the satellite image sequence can possess subjective user-specific semantics (e.g.  harvests, wheat harvests or crop changes in general). A user may be interested in retrieving similar events and thus, may want to know when and where similar spatio-temporal patterns have occurred. An example of probabilistic retrieval of spatio-temporal patterns occurring in  an image sequence according to a user semantic is given in Fig.~\ref{fig1}. Moreover, as graph patterns $\mathcal{G}_k$ contained in $\mathcal{G}$ are stochastic models for these spatio-temporal patterns, they can also possess a user semantic. Therefore, we are interested in learning a semantic from a user in order to achieve a semantic labeling of graph patterns representing spatio-temporal patterns which enables the recognition and the probabilistic retrieval of similar spatio-temporal phenomena.\\

\noindent
Until now, learning methods for satellite image sequences have been dedicated to the analysis and recognition of particular spatio-temporal phenomena in relation to applications such as change detection~\cite{Bruzzone(2000)}, data assimilation for agriculture monitoring~\cite{Lauvernet(2003)} or wind field extraction~\cite{Corpetti(2002)}. Although these techniques are efficient, together they represent a limited range of applications. Until now, only few methods mainly focusing on low resolution images regularly sampled in time~\cite{Antunes}\cite{Tan}\cite{Nasa} have been developed in order to adapt to a broader range of application. However, to access to the variety of information contained in high resolution satellite image sequences, collaborative and generic methods are needed.\\

\noindent
In this paper we propose an original learning method responding to this problematic. The remainder of the paper is organized as follows. After a description of the global supervised semantic modeling procedure,  we present the parametric model used for evaluating similarity between graph patterns. Then, we propose a Bayesian approach for learning the distribution of the similarity parameters based on a Dirichlet model and user-provided examples. The learning process yields to the estimation and the semantic labeling stages. Finally, after a section describing experimental results, a short summary concludes the discussion.

\section{Bayesian modeling of user semantics}

The inference of the graph $\mathcal{G}$ is a robust and unsupervised coding of satellite image sequences. Based on this objective signal characterization, we focus now on modeling by user-provided examples the semantics attached to spatio-temporal patterns in satellite image sequences. The proposed supervised learning approach is based on Bayesian networks~\cite{Cooper(1992)}\cite{Heckerman(1999)}. It aims in extending the learning system proposed in~\cite{Schroder(2000)} to spatio-temporal features.\\

\noindent
In order to define a model for a given user semantic $\mathcal{A}_\nu$, we introduce a parametric similarity cost $S_\Phi(\mathcal{G}_0,\mathcal{G}_k)$ between the graph pattern $\mathcal{G}_k$ and a reference graph pattern $\mathcal{G}_0$. Dynamic time warping schemes~\cite{Berndt(1996)} constitute efficient approaches for evaluating graph pattern similarities. However, the extension of such a distance measurement to multidimensional graph features of heterogeneous nature is not obvious. A simple solution has been chosen here to deal with such multidimensional graph patterns. We build a parametrical model for similarity by extending the inexact graph matching algorithm proposed in~\cite{Bunke(1983)}. In the introduced model, a parameter vector denoted by $\Phi$  weights the contribution of each type of graph features.  This parametrical model will be detailed in section~\ref{ParamModel}.\\

\noindent
An intuitive assumption is that a given parameter vector corresponds to a particular similarity, which can formalize a given user semantic. Therefore, parameters can be tuned in order to represent a given user semantic. We will see in section~\ref{intLearning}, that parameters $\Phi$ of the similarity model and the reference graph $\mathcal{G}_0$ can be estimated via an supervised learning process relying on user-provided examples. It is  thus possible to link subjective elements $\mathcal{A}_\nu$ representing user semantics to graph patterns $\mathcal{G}_k$. In this perspective, we make  the hypothesis that a parametric similarity cost $S_\Phi(\mathcal{G}_0,\mathcal{G}_k)$ constitutes a model $\mathcal{M}$ which is sufficient for describing the different semantics. And, introducing a normalization constant $Z$, we define simply the likelihood probability of the semantic $\mathcal{A}_\nu$  for each graph pattern $\mathcal{G}_k$ as :
\begin{eqnarray}\label{likelihood}
p(\mathcal{G}_k \mid \mathcal{A}_\nu,\mathcal{M})=1-\frac{ S_{\widehat{\Phi}}(\widehat{\mathcal{G}}_0,\mathcal{G}_k)}{Z},
\end{eqnarray}
where  $\widehat{\Phi}$ and $\widehat{\mathcal{G}}_0$ are respectively a parameter vector and a reference graph, both estimated via learning with examples.  For notation simplification, the conditioning of the likelihood by a model $\mathcal{M}$ is omitted in the following.\\

\noindent
Based on these likelihood probabilities, using a Bayesian context enables the estimation of posterior probabilities $p(\mathcal{A}_\nu \mid \mathcal{G}_k)$ and thus, allows a semantic representation of the satellite image sequences content.
Indeed, considering that a user provides positive and negative examples, corresponding to a positive $\mathcal{A}_\nu$ and a negative $\neg \mathcal{A}_\nu$ semantic,  two likelihood probabilities $p(\mathcal{G}_k \mid \mathcal{A}_\nu)$ and $p(\mathcal{G}_k \mid \neg \mathcal{A}_\nu)$ can be derived for each graph patterns. Moreover, graph priors can be obtained using the formula  $p(\mathcal{G}_k)=\sum_{i}p(\mathcal{G}_k \mid \mathcal{A}_i)p(\mathcal{A}_i)$, where the summation is done over the positive $\mathcal{A}_\nu$ and negative $\neg \mathcal{A}_\nu$ semantics. Thus, assuming a uniform prior on the semantics, the posterior probabilities of the positive semantic are inferred using Bayes rule :
\begin{eqnarray}\label{Bayes}
p(\mathcal{A}_\nu \mid \mathcal{G}_k)&=&\frac{p(\mathcal{G}_k \mid \mathcal{A}_\nu)p(\mathcal{A}_\nu)}{p(\mathcal{G}_k)}\nonumber \\
&=&\frac{p(\mathcal{G}_k \mid \mathcal{A}_\nu)}{p(\mathcal{G}_k \mid \mathcal{A}_\nu)+p(\mathcal{G}_k \mid \neg \mathcal{A}_\nu)}.
\end{eqnarray}
Thus, to achieve the posterior estimation, we need to define : (1) a parametric cost $S_{\widehat{\Phi}}(\widehat{\mathcal{G}}_0,\mathcal{G}_k)$ for graph pattern similarity to enable the evaluation of likelihood probabilities $p(\mathcal{G}_k \mid \mathcal{A}_\nu)$ and $p(\mathcal{G}_k \mid \neg \mathcal{A}_\nu)$, (2) a method for learning by examples the model parameters $\widehat{\Phi}$ and $\widehat{\mathcal{G}}_0$ needed for the evaluation of  the previous likelihood probabilities. These points are detailed in the two next sections.\\

\section{Parametric model evaluating graph pattern similarity}\label{ParamModel}

The idea of inexact graph matching is to transform one of the graph patterns into the other one by assigning a cost to each vertex or edge addition/removal. 
However, graph patterns $\mathcal{G}_k$ are specific multidimensional temporal features which characterize parts of the dynamic cluster trajectories. More precisely, they correspond to given classes of a multitemporal classification within a given temporal window. The information is condensed in vertices and edges. A vertex is representing a multivariate Gaussian distribution related to a given spatial class at a given time. It is characterized by a pixel weight, Gaussian parameters and a divergence measurement which has been used for the trajectory reconstruction. An edge, representing the evolution of the cluster between two image samples, is characterized by a time sampling delay, a pixel flow, Gaussian parameter evolution and multitemporal intra-class changes quantified by mutual information\footnote{For more details on the trajectory attributes please refer to~\cite{Heas(2005)}}. Let us denote by $\{\zeta_l\}$ the set of attributes related to a graph patterns. Thus, the inexact graph matching algorithm is extended to a parametric distance model between graph patterns, weighting the different attribute contributions.\\

Denoting by $\nu_1=\{\nu^1_i\}$ and $\nu_2=\{\nu^2_i\}$ the vertex sets of graph patterns $\mathcal{G}_1$ and $\mathcal{G}_2$, and denoting  an extra set of vertices  by $\lambda=\{\lambda_i\}$, a mapping function $\mathcal{F}=\{f\}$ composed by a given combination of elementary mapping functions $f:\nu^1 \rightarrow \nu^{2\lambda}=\nu^2\cup \lambda$  is defined. A cost $C_\Phi(f(\nu^1_i)=\nu^{2\lambda}_j)$ is assigned to each elementary transformations. The cost function depends on the  parameter vector $\Phi=\{\phi_l\}$ and is composed by a weighted sum of similarities between vertices $\nu^1_i$ and $\nu^{2\lambda}_j$ and related edges.  The cost is equal to
\begin{eqnarray}\label{sommeAtt}
C_\Phi(f(\nu^1_i)=\nu^{2\lambda}_j)= \sum_l \phi_l\Delta_l(\zeta_l(\nu^1_i),\zeta_l(\nu^{2\lambda}_j)) 
\end{eqnarray} 
where $\Delta_l(.)$ represents a distance model which is either a difference for scalars or a similarity cost between probability density functions such as Kullbach-Leibler divergence.
The graph patterns similarity is then defined, for a given vector parameter $\Phi$, by finding the less expensive elementary mapping function combination over all possible mapping functions:
\begin{eqnarray}\label{distance}
S_\Phi(\mathcal{G}_1,\mathcal{G}_2)=\min_{\mathcal{F}}\big( \sum_{i} C_\Phi(f(\nu^1_i)=\nu^{2\lambda}_j)\big).
\end{eqnarray} 
Denoting by $S_{l}(\mathcal{G}_1,\mathcal{G}_2)$ the cost related to parameter $\phi_l$ in the similarity function $S_\Phi(\mathcal{G}_1,\mathcal{G}_2)$, Eq.~\ref{distance} is rewritten as
\begin{eqnarray}
S_\Phi(\mathcal{G}_1,\mathcal{G}_2)&=&\sum_{l} \phi_l\min_{\mathcal{F}}\big( \sum_{i}\Delta_l(\zeta_l(\nu^1_i),\zeta_l(\nu^{2\lambda}_j)) \big) \nonumber \\
&=&\sum_{l}\phi_l S_{l}(\mathcal{G}_1,\mathcal{G}_2).
\end{eqnarray} 
In order to estimate the minima, an optimization procedure is performed searching a minimum cost path in a tree containing all possible mapping functions configurations. Because, of the combinatorial explosion of configurations and real-time requirements, the tree is pruned during the search according to the current cost assigned to the branches. This optimization procedure is obviously sub-optimal for dense graph patterns with the potential drawback of yielding to local minima. Thus, the pruning approach constitutes an easy solution for matching simple graph patterns i.e. with few vertices and edges. However, we remark that optimization strategy based for example on graph-cuts~\cite{Boykov(2001)} should be considered for more complex graph patterns.

\section{Learning the similarity model parameters}\label{intLearning}
  
In the previous section, we developed a similarity cost function between graph patterns which depends on a parameter vector $\Phi$. The different components of this vector weight the different contributions related to graph attributes $\zeta_l$ composing the global similarity cost $S_\Phi(.)$. As it has already been mentioned, we make the assumption that a given parameter vector corresponds to a particular similarity which can formalize a semantic related to a user. But the manual tuning of the parameters in order to define a similarity specific to a semantic may represent a tedious task or even an impossible task for a user.  Therefore, a supervised learning procedure is needed to estimate the parameter vector $\Phi$, enabling via similarity costs, the evaluation of semantic likelihoods $p(\mathcal{G}_k \mid \mathcal{A}_\nu)$ and $p(\mathcal{G}_k \mid \neg \mathcal{A}_\nu)$, which are then used for the inference of posterior probabilities $p(\mathcal{A}_\nu \mid \mathcal{G}_k)$.\\
We detail in the following how the parameter distribution related to the positive semantic likelihood is learned by user-provided examples and how the parameter estimation process is performed. Parameters related to the negative semantic likelihood are obtained in a similar framework. Finally, learning result in the semantic labeling of the different graph patterns present in the satellite image sequence.

\subsection{Multinomial models for discretized parameter distributions}

The idea for the supervised estimation of the similarity model parameters according to a given semantic is the following~: we consider a given reference graph pattern $\mathcal{G}_0$ and an example provided by the user of a spatio-temporal phenomenon (i.e. a graph pattern $\mathcal{G}_k$) which possesses a given semantic $\mathcal{A}_\nu$; then, the lower the partial cost $S_{l}(\mathcal{G}_0,\mathcal{G}_k)$  related to the attribute $\zeta_l$, the more important the weight $\phi_l$. In other words, we make the assumption that the cost function $S_{l}(\mathcal{G}_0,\mathcal{G}_k)$ related to the attribute $\zeta_l$ is proportional to the opposite of the parameter value $\phi_l$ :
\begin{eqnarray}
\phi_l \propto -S_{l}(\mathcal{G}_0,\mathcal{G}_k).
\end{eqnarray}
Let us now take advantage of the previous  proportionality  assumption. First, to allow a comparison between the different parameters $\phi_l$, we normalize the  domain where the cost functions $S_{l}(\mathcal{G}_0,\mathcal{G}_k)$ take their values. Then, as the estimation of a continuous distribution is difficult when very little data is available, the continuous parameters $\{\phi_l\}$ are discretized in $r$ quantization levels, so that each parameter $\phi_l$  take their values in $\{\phi^1_l,...,\phi^r_l\}$ and follow a multinomial law\footnote{The number of quantization level $r$ should be sufficiently large in order to approximate a continuous distribution. This number $r$ should also be chosen according to the number of examples provided by the user during the learning process. In this work, $r$ was fixed to 1000.}. The latter distribution has the advantage of possessing parameters linked to occurrence probabilities, which as we will see, can be estimated in real time in a Bayesian context.\\  

\noindent
Thus, considering the user semantic  $\mathcal{A}_\nu$, the conditioned probability density function is defined for $j=1,...,r$ by
\begin{eqnarray}\label{multinomial}
p(\phi_l=\phi_l^j \mid \omega, \mathcal{A}_\nu)&=&p(\Lambda (S_{l}(\mathcal{G}_0,\mathcal{G}_k))=\phi_l^j \mid \omega, \mathcal{A}_\nu) \nonumber \\
&=&\omega_j,
\end{eqnarray}
where $\omega=\{\omega_2,...,\omega_r\}$ are the parameters of the multinomial model\footnote{Note that parameter $\omega_1$ is given by $1-\sum_{j=2}^{r}\omega_j$} and $\Lambda(.)$ is an operator discretizing the normalized interval where the functions $S_{l}(\mathcal{G}_0,\mathcal{G}_k)$ take their values, in $r$ quantization levels $\{\phi^1_l,...,\phi^r_l\}$. For notation simplifications, $p_{\mathcal{G}_0}(\phi_l=\phi_l^j \mid \omega, \mathcal{A}_\nu)$ will be noted $p(\phi_l^j \mid \omega, \mathcal{A}_\nu)$.\\ 

\noindent
Furthermore, statistical independence is assumed on the parameter conditioned distribution in order to avoid the joint probability distribution estimation. 
Note that this assumption is necessary to reduce the model complexity and thus, allow the interactive learning which will be presented in the following. However, the validity of such an assumption depends on the nature of graph pattern attributes $\zeta_l$ used in the similarity model. For example, pixel flows can be assumed independent from Gaussian parameters. On the contrary, mutual information is not necessary independent from pixel flows. Nevertheless, assuming the latter assumption valid, we obtain~:\\
\begin{eqnarray}p(\Phi \mid \mathcal{A}_\nu)=p(\phi_{1} \mid \mathcal{A}_\nu)p(\phi_{2} \mid \mathcal{A}_\nu)\,...
\end{eqnarray} 

\subsection{Supervised learning of multinomial distributions}

For a given semantic $\mathcal{A}_\nu$, we now move the discussion from assessing the probability $p(\phi_l \mid \mathcal{A}_\nu)$ of each parameter $\phi_{l}$, to assessing the probability distribution $p(\omega \mid \xi)$ of parameters $\omega$ attached to the multinomial model, where $\xi$ denotes a given level of knowledge.\\ 

\noindent
Supervised learning proposed in this section is inspired of previous work on learning with Bayesian networks~\cite{Heckerman(1999)}\cite{Schroder(2000)}. 
Learning is performed via training the system by a user.  A Bayesian framework is adopted because of its robustness when very limited user examples are available. The user provide a training dataset $T$ of graph patterns examples in accordance to his semantic. With those user-provided examples, we define for each parameter $\phi_l$, a vector   $N=\{N_1,...,N_r\}$ with $N_j$ being the number of instance of $\phi_l^j$, that is the number of times that $\phi_l=\phi_l^j$ occurs in examples $T$. Note that parameters $\omega$ of the multinomial distribution correspond to occurrence probabilities.\\

\noindent
For the supervised evaluation of the occurrence probabilities (or the multinomial model parameters), we introduce  the  Dirichlet distribution as a conjugate prior. For a given level of knowledge $\xi$, this distribution depends on a vector of hyper-parameters $\alpha=\{\alpha_1,....,\alpha_r\}$ and is expressed by   
\begin{eqnarray}
p(\omega \mid \xi)&=&Dir(\omega \mid \alpha_1,...,\alpha_r)\nonumber \\
&=&\frac{\Gamma(\alpha)}{\prod_{k=1}^r \Gamma (\alpha_k)} \prod_{j=1}^r (\omega_j)^{\alpha_j-1}
\end{eqnarray}
 where $\alpha=\sum_{j=1}^r\alpha_j$ and $\alpha_j > 0, \forall j \in[1,r]$ and where $\Gamma(x)$ denotes the Gamma function.\\

\noindent
The learning of a multinomial distribution (Eq.~\ref{multinomial}) uses for initialization, the Dirichlet function with all hyper-parameters $\alpha_j$ equal to one which represents a uniform probability density function.
The prior Dirichlet function is 
\begin{eqnarray}
p(\omega)=Dir(\omega \mid \alpha_1^{(0)},...,\alpha_r^{(0)}); \forall j \in [1,r] ,\alpha_j^{(0)}=1.
\end{eqnarray}
After observing the instances $\{N_j^{(1)}\}$ in a training dataset $T^{(1)}$, according to Bayes rule, the posterior probability is
\begin{eqnarray}
p(\omega \mid T^{(1)})&=&\frac{p(T ^{(1)}\mid \omega)p(\omega)}{p(T^{(1)})}  \\
&=&Dir(\omega \mid \alpha_1^{(0)}+N_1^{(1)},...,\alpha_r^{(0)}+N_r^{(1)}) \nonumber
\end{eqnarray}
After observing another training dataset $T^{(2)}$, which is assumed to be independent from $T^{(1)}$ we obtain the new posterior 
\begin{eqnarray}
p(\omega \mid T^{(2)},T^{(1)})&=&\frac{p(T^{(2)} \mid \omega,T^{(1)})p(\omega \mid T^{(1)})}{p(T^{(2)})}  \\
&=&Dir(\omega \mid \alpha_1^{(1)}+N_1^{(2)},...,\alpha_r^{(1)}+N_r^{(2)}) \nonumber
\end{eqnarray}
where the new hyper-parameters were calculated by adding the number of times $\phi_l^{j}$ occurred in the training data set $T^{(2)}$.
Therefore, each observed set of data $T^{(i)}$ can be incorporated as an update of the hyper-parameters : $\alpha_j^{(i)}=\alpha_j^{(i-1)}+N_j^{(i)}$.\\

\noindent
Considering some training $T$ with the associated hyper-parameter vector $\alpha$, the estimation of $p(\phi_l^j \mid  \mathcal{A}_\nu, T)$ is achieved using the Minimum Mean Square Error (MMSE) estimator of the parameter $\omega_j$ :
\begin{eqnarray}
p(\phi_l=\phi_l^j \mid  \mathcal{A}_\nu)&=&E[\omega_j]=\int \omega_j p(\omega \mid T) d\omega \nonumber \\
 &=& \frac{\alpha_j}{\alpha}.
\end{eqnarray}
Finally, by using the independence assumption, we obtain $p(\Phi \mid \mathcal{A}_\nu)$ by making the product $\prod_l p(\phi_l \mid  \mathcal{A}_\nu)$.

\subsection{Estimation and update of the similarity model parameters}

After some training $T$, one can use the MMSE estimator to evaluate the parameter vector $\Phi$ of the similarity function. It is defined by 
\begin{eqnarray}
\widehat{\Phi}_{MMSE}=E[\Phi],
\end{eqnarray}
 where $E[.]$ is the expectation operator related to the probability distribution $p(\Phi \mid \mathcal{A}_\nu)$. We note that the multinomial distribution does not show a clear maximum because of the too few examples provided by the user compared to the large number $r$ of values $\phi_l^j; l=1,...,r$. This justifies the use of the MMSE estimator rather than the  maximum a posteriori estimator.\\

\noindent
Using this parameter vector update, we perform a new evaluation of the similarity function. Therefore, according to Eq.~\ref{likelihood}, a semantic likelihood probability can be assigned to each graph pattern  $\mathcal{G}_k$  with this new estimate :
\begin{eqnarray}
p(\mathcal{G}_k \mid \mathcal{A}_\nu)=1-\frac{ S_{\widehat{\Phi}_{MMSE}}(\widehat{\mathcal{G}}_0,\mathcal{G}_k)}{Z}.
\end{eqnarray}
We choose to use a uniform distribution to initialize the parameter vector distribution.\\

\noindent
Note that the latter probabilities are dependent of an estimated reference graph $\widehat{\mathcal{G}}_0$. Using Eq.~\ref{Bayes},  enables the evaluation of likelihood probabilities $p(\mathcal{G}_k \mid \mathcal{A}_\nu )$  for each graph pattern $\mathcal{G}_k$. We thus obtain a new estimate for the reference graph pattern by selecting the one which maximizes the likelihood probability :
\begin{eqnarray}
\widehat{\mathcal{G}}_0=arg \max_{\mathcal{G}_k} p(\mathcal{G}_k \mid \mathcal{A}_\nu ).
\end{eqnarray}  
The first example provided by the user determines the initial reference graph pattern. It is then updated according to the previous equation after each learning iteration. The reference graph pattern related to the negative semantic is initialized and updated similarly.

\subsection{Semantic labeling of graph patterns}

\begin{figure*}[h!]
\begin{center}
\begin{tabular}{c}
\includegraphics[width=1.\textwidth]{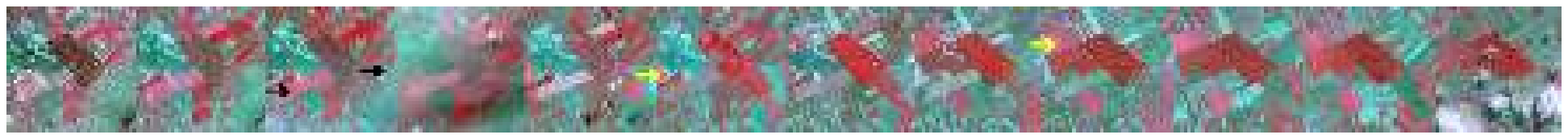}\\
\includegraphics[width=0.55\textwidth]{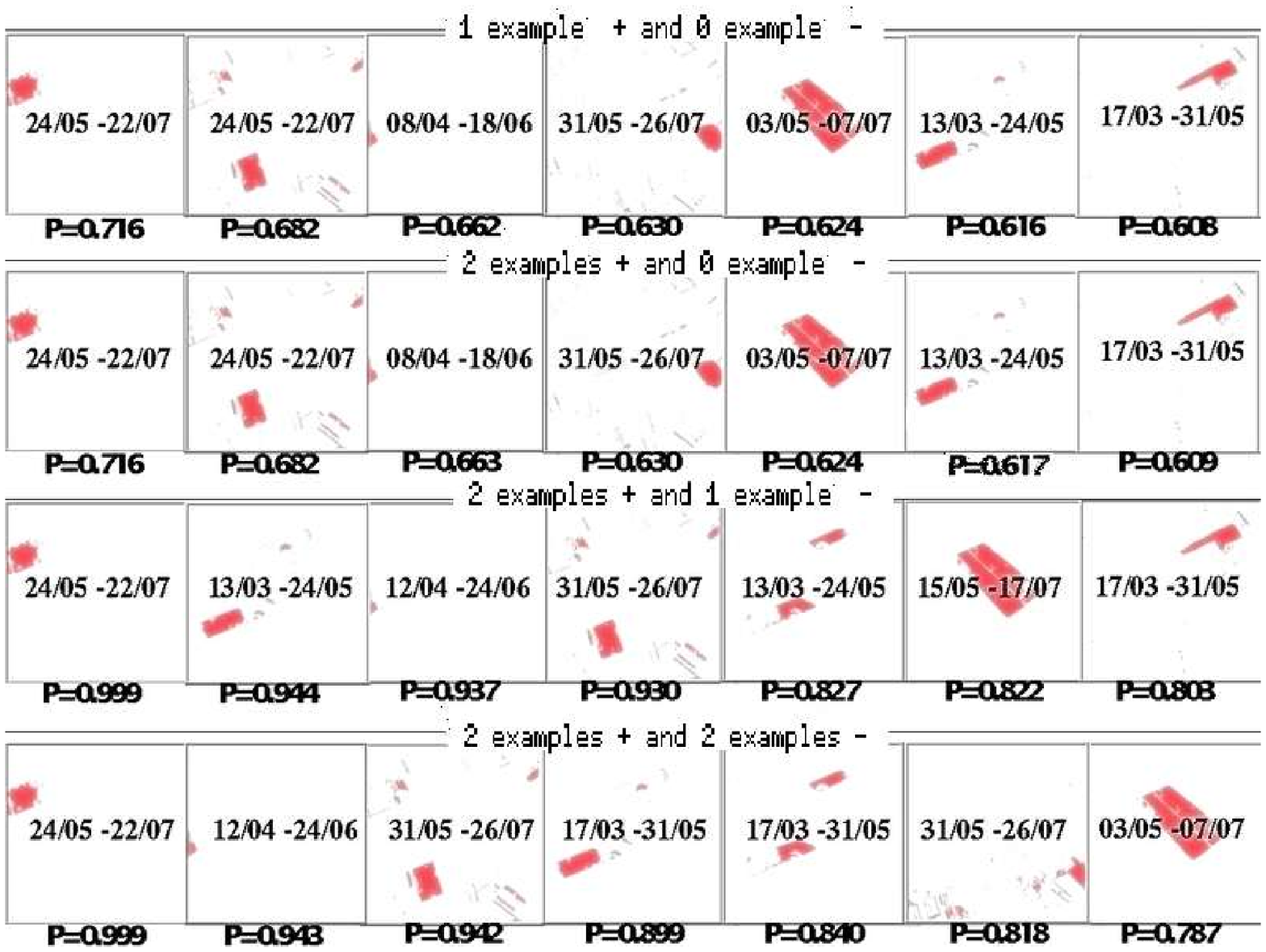}
\end{tabular}
\caption{ \textit{\small{Iterations for learning field maturation semantics. \textbf{Above} : for visualization purposes, the image sequence has been subsampled temporally, and only some of the 38 images are here displayed;  two pattern examples related to a positive semantic (yellow arrows) and two other examples related to a negative semantic (black arrows) are  successively  introduced; the arrows designing these pattern examples are represented with arrows indicating fields in the apogee of their maturation process within a temporal window of 12 time samples. \textbf{Bellow}~: collections of spatio-temporal patterns possessing the highest posterior probabilities $P$ retrieved after each example provided by the user; each line represents the current collection of spatio-temporal patterns retrieved which possess the highest probabilities; these patterns are defined by spatial classes displayed in red and by the temporal windows indicated at the center of the classes.}}\label{Training}}
\end{center}
\end{figure*}

\begin{figure}[!h]
\centering
\begin{tabular}{c}
\includegraphics[width=0.5\textwidth]{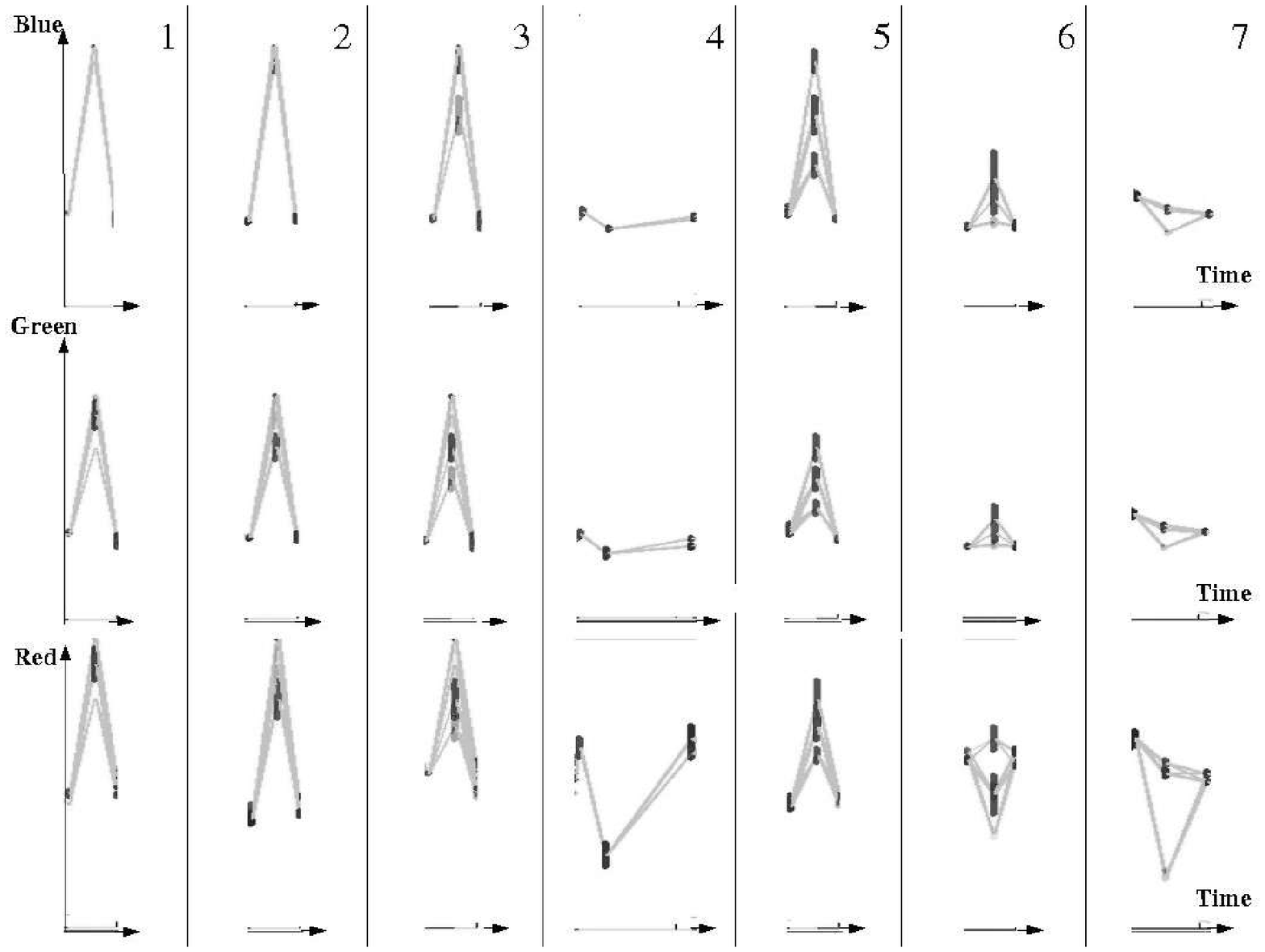}\\
\includegraphics[width=0.5\textwidth]{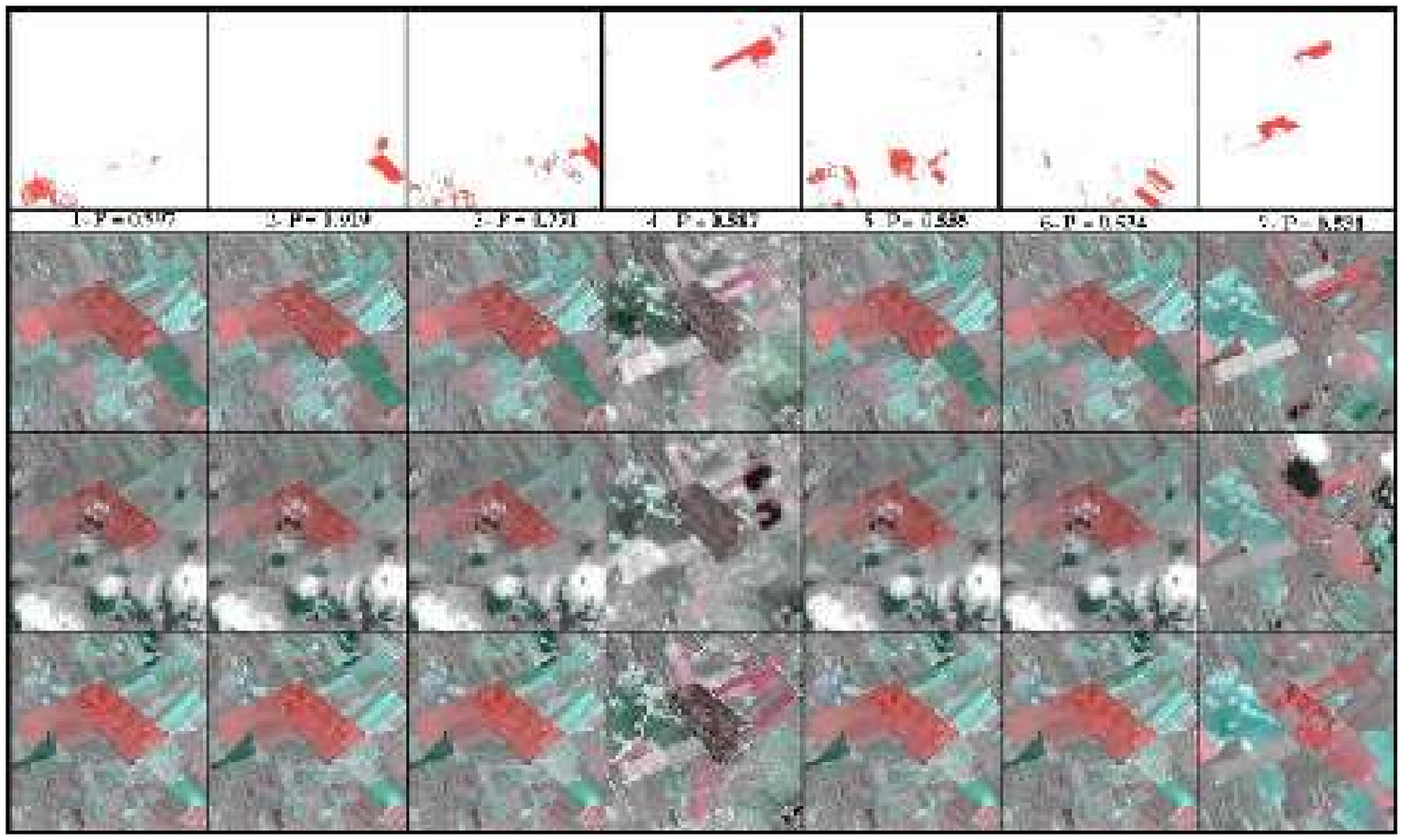}
\end{tabular}
\caption{Retrieval of clouds based on morphological similarities of graph patterns. Each column represents a spatio-temporal phenomenon which has been probabilistically labeled with a cloud occlusion semantic. The phenomena are represented above in the form of graph patterns, that is to say projections of parts of cluster trajectories (temporal window of 3 time samples) living in the  3D spectral feature space (Red-Green-Blue). In the middle, for each column, the corresponding  spatial class where the phenomenon occurred is displayed in red with its posterior probability $P$. Bellow, for each column, 3 time samples of the image sequence comprising the corresponding spatio-temporal pattern (i.e. cloud occlusion) is displayed.}  
\label{clouds}
\end{figure}

In the previous sections, the learning of the  positive and negative semantic likelihood probabilities has been detailed. Using the Bayesian semantic modeling of Eq.~\ref{Bayes} yields to the update of posterior probabilities $p( \mathcal{A}_\nu \mid \mathcal{G}_k)$ for each graph pattern $\mathcal{G}_k$ after each examples of spatio-temporal phenomena provided by the user.
An example showing the successive probability updates resulting from supervised learning of field maturation semantics is presented for sake of clarity in figure~\ref{Training}. For the visual inspection of graph morphological similarities which have been learned after training a cloud occlusion semantic,  in figure~\ref{clouds} we have plotted graph patterns possessing high posterior probabilities with their associated spatio-temporal phenomena. Note that very few clouds are remaining in the image sequence as image with large cloud coverage were previously filtered out. In the two latter examples, the features were the 3 spectral reflectances extracted out of the image sequence in  a spatial subset of 200x200 pixels.\\

\noindent
We rely on those posterior probabilities to attach to the graph patterns, i.e. to the spatio-temporal phenomena, semantic labels. We consider that a phenomena possesses a semantic label $\mathcal{A}_\nu$ if the posterior probability exceeds a false alarm threshold chosen by the user.

\section{Experiments}

\begin{figure}[!h]
\centering
\includegraphics[width=3.1in]{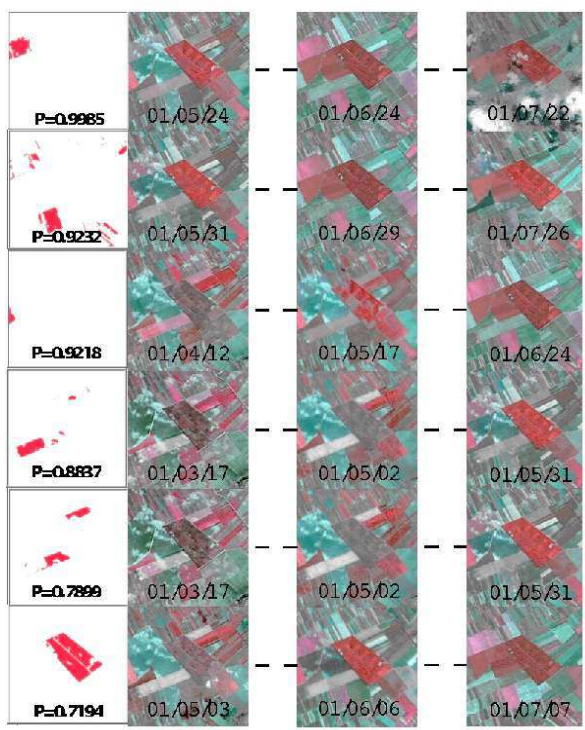}
\caption{Supervised learning of field maturation semantics : most likely  spatio-temporal structures retrieved in a spatial window of 200x200 pixels and ranked, from top to bottom, according to their posterior probabilities. Each row presents a retrieved spatial class (left) with its associated time-period, which is given by time locations in the first and last images of the row. The middle images in each row correspond to the time sample (within the temporal window of 12 samples) where maturations reached their apogees.}
\label{maturation}
\end{figure}

In the experiments carried out, we first focused on spectral features extracted out of the image sequence in a spatial subset of 200x200 pixels. We trained maturation semantics, specific to a field. As these phenomena occurred over  a long time period, a time window of 12 samples was selected for training. With very few positive and negative examples, the supervised learning process enabled the retrieval of similar events with high posterior probabilities. The retrieved spatio-temporal structures are presented in figure~\ref{maturation} together with 3 significant image time samples. Note that, the crop evolutions of highest probabilities are maturation phenomena corresponding to the specific sought culture, whereas retrieved events with lower probabilities correspond to maturation of similar but slightly different cultures.\\

\begin{figure*}[!h]
\begin{center}
\begin{tabular}{cc}
\hspace{5.5cm}\includegraphics[width=0.32\textwidth]{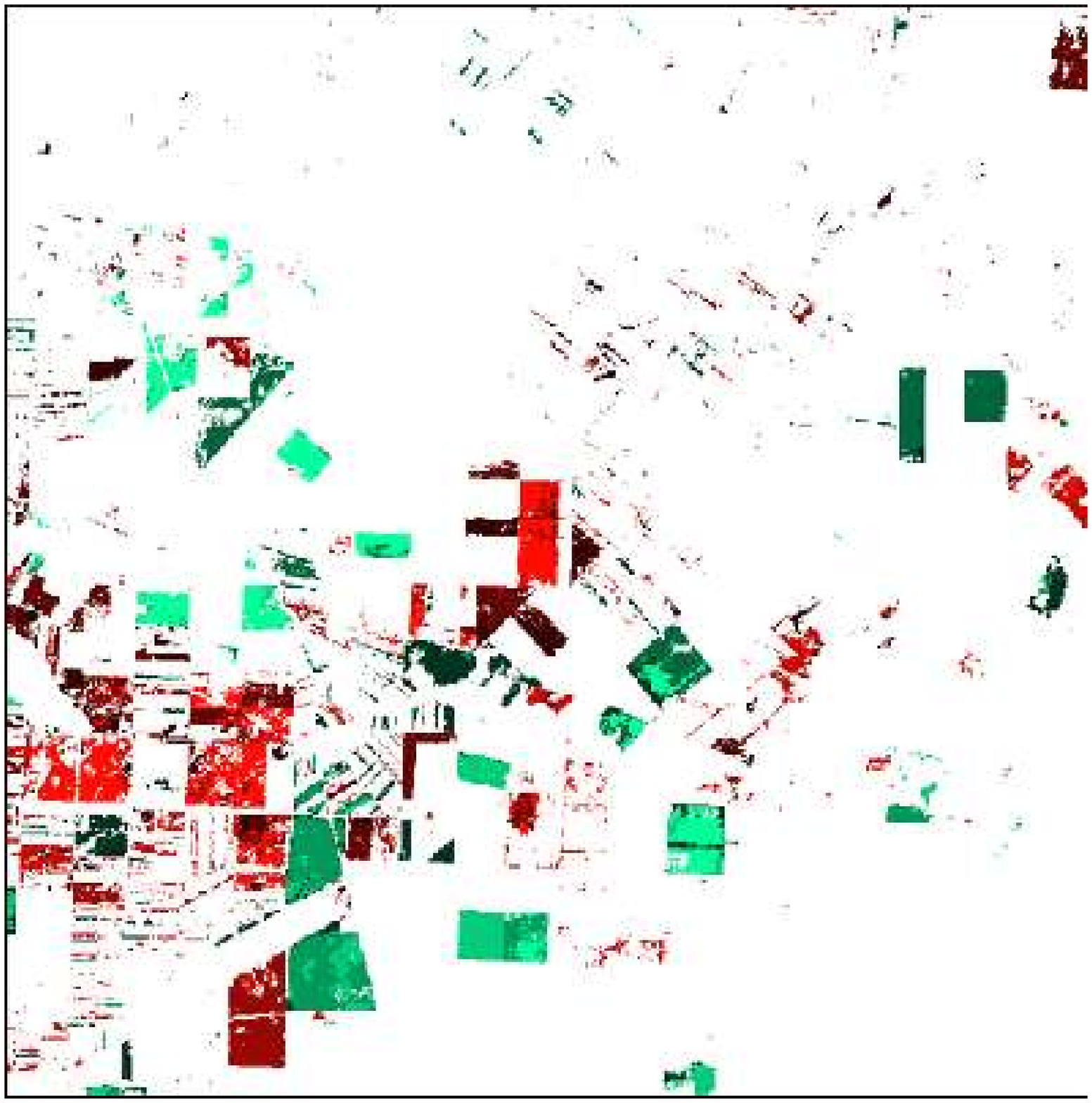}&\hspace{-0.1cm}\includegraphics[width=0.2\textwidth]{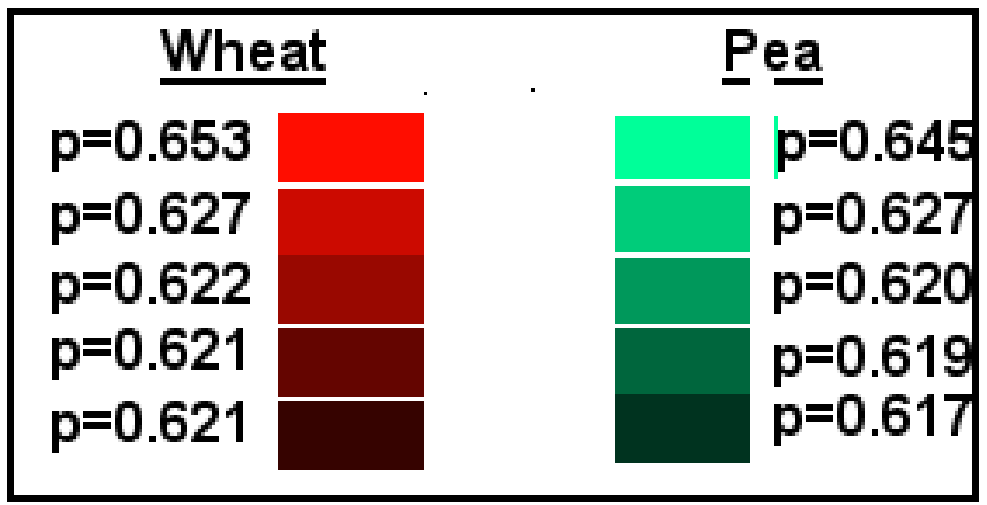}\\
\multicolumn{2}{c}{\includegraphics[width=1.0\textwidth]{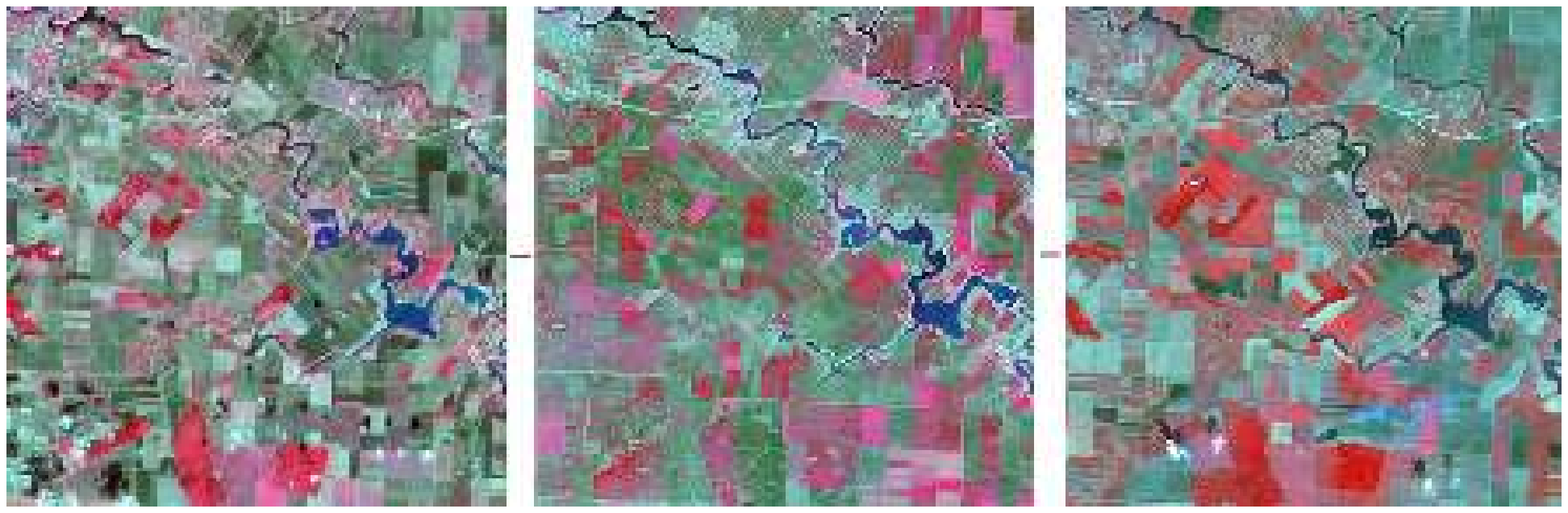}}
\end{tabular}
\caption{ \textit{\small{Recognition of particular farming practices. Crops of wheat or peas related to particular farming practices were retrieved by supervised learning in a spatial window of 800x800 pixels. Similar spatio-temporal structures defined within a maximum temporal window constituted by 38 time samples of the sequence where retrieved in space. A single example of evolution related to a crop of wheat or peas enabled the recognition of fields undergoing similar farming practices (same harvest period, plowing, etc). \textbf{Above}~: for both farming practices (wheat or pea), retrieved multitemporal classes are displayed with shaded colors (red or green) according to their posterior probabilities $p$ appearing in the caption on the right. \textbf{Bellow}~: for visualization purposes, the image sequence is represented here by the first  image (on the left), the last  image (on the right) and a single intermediate image (in the middle).}}} 
\label{blePois}
\end{center}
\end{figure*}

\noindent
Experiments were then performed with spectral features extracted out of the image sequence in a spatial subset of 800x800 pixels. A search was launch to identify crops undergoing similar farming practices within 286 days of observations, that is to say the whole image sequence. We particularly focused on the wheat annual farming cycle : in autumn, crops are plowed and then sowed with wheat; the crop vegetates during winter and in spring the plants grow up to maturation; at the end of summer the wheat is finally harvested. We also identified pea farming : the evolution is characterized by the development of leaves and ramifications in spring, a flowering in the beginning of June and a harvest in August. Therefore, a single example of a crop of wheat or peas undergoing such a farming process was provided to the system for training.\\
Results are displayed in figure~\ref{blePois}. In order to understand why the greatest posterior probabilities have been attached to those structures, a careful inspection of the image sequence was performed. This laborious task enabled us for example to identify similar crops which have not been retrieved because of an early harvest.
Let us remark that the repartition of the classes is quite sparse. Therefore, this example demonstrates the capacity of the proposed learning approach to recognize complexes phenomena, spread in space and undergoing similar changes in time. Note that achieving a similar task by visual inspection would have been considerably time-consuming.\\

\noindent
Let us also mention the limitations induced by the graph matching optimization algorithm which has been used in the latter experiments. Selecting a limited spatial window (first experiments) or defining searched phenomena within a maximum temporal window (last experiment) reduces considerably the number of graph pattern $\mathcal{G}_k$ contained in the whole graph $\mathcal{G}$. Moreover, in the previous experiments spatio-temporal phenomena have been coded with simple graph patterns\footnote{For more details on tuning graph pattern complexity, please refer to~\cite{Heas(2005)}}. Therefore, the calculation of graph likelihoods  has been performed in real time and posterior probabilities have appeared to be relevant of the different user semantics. Nevertheless, learning semantics attached to numerous spatio-temporal phenomena coded with too dense graph patterns may be time-consuming and not respect real time requirements. Indeed, the combinatory explosion problem for matching vertices and edges is accentuated for dense graph patterns. Thus, our simple optimization algorithm may not reveal  sufficiently accurate parameter  minima and result in a weak learning. Thus, implementing a better optimization algorithm based for example on graph cuts is required for further experiments.

\section{Conclusion and perspectives}
This work is an attempt to solve the complex problem of recognizing various spatio-temporal phenomena in satellite image sequences. The proposed concept, developed in a Bayesian framework, models a user semantic by  a parametric model evaluating the similarities of graph patterns. The latter code spatio-temporal phenomena. Discretized parameter distributions related to the similarity model are learned in a supervised way by updates of the parameters of multinomial models. The learning process is based on a Dirichlet model and user-provided examples related to positive and negative semantics.\\   
Based on results on SPOT image sequence, the method appears to be a fast and relevant way to retrieve user-specific spatio-temporal patterns. The experiments have also revealed that the optimization algorithm used for evaluating graph pattern similarity constitutes a crucial issue for further developing the learning capabilities.\\

\noindent
We believe that the learning concept we have presented constitutes a valuable tool in view of the numerous potential applications. Collecting ground truth data or available expert knowledge related to agriculture or other applications will be the next step towards the exhaustive assessment of the proposed spatio-temporal recognition approach. Moreover, such a supervised learning method can apply on  multidimensional graph coding any data. Using this approach in other fields, such as molecular biology or telecommunication networks for the recognition of particular graph patterns, constitute a very interesting perspective. 

{\small 
\bibliographystyle{splncs}

\end{document}